%% file: main.tex
\documentclass{article}

\usepackage{microtype}
\usepackage{graphicx}
\usepackage{subcaption}
\usepackage{booktabs}

\usepackage{hyperref}

\usepackage[accepted]{icml2026}

\usepackage{amsmath}
\usepackage{amssymb}
\usepackage{amsfonts}
\usepackage{mathtools}
\usepackage{amsthm}

\usepackage{multirow}
\usepackage{xcolor}
\usepackage{colortbl}
\usepackage{enumitem}
\usepackage{array}
\usepackage{tabularx}
\usepackage{makecell}
\usepackage{placeins}
\usepackage{float}
\usepackage{needspace}

\usepackage[capitalize,noabbrev]{cleveref}

\newcolumntype{C}[1]{>{\centering\arraybackslash}p{#1}}
\newcolumntype{R}[1]{>{\raggedleft\arraybackslash}p{#1}}

\definecolor{bestcolor}{HTML}{D4EDDA}
\definecolor{worstcolor}{HTML}{F8D7DA}
\definecolor{lightgray}{gray}{0.92}

\icmltitlerunning{DistortBench: Benchmarking Vision Language Models on Image Distortion Identification}

\begin{document}

\twocolumn[
  \icmltitle{DistortBench: Benchmarking Vision Language Models on Image Distortion Identification}


  \icmlsetsymbol{equal}{*}

  \begin{icmlauthorlist}
    \icmlauthor{Divyanshu Goyal}{adobe}
    \icmlauthor{Akhil Eppa}{adobe}
    \icmlauthor{Vanya Bannihatti Kumar}{adobe}
  \end{icmlauthorlist}

  \icmlaffiliation{adobe}{Adobe Inc., San Jose, CA, USA}

  \icmlcorrespondingauthor{Divyanshu Goyal}{divgoyal@adobe.com}
  \icmlcorrespondingauthor{Akhil Eppa}{aeppa@adobe.com}
  \icmlcorrespondingauthor{Vanya Bannihatti Kumar}{vbannihatti@adobe.com}

  \icmlkeywords{Vision-Language Models, Image Quality Assessment, Benchmark, Distortion Identification, Perceptual Quality}

  \vskip 0.3in
]

\printAffiliationsAndNotice{Presented as a spotlight
at the Efficient Multimodal Question Answering
(EMM-QA) Workshop at ICML 2026.}

\input{sec/0_abstract}
\input{sec/1_intro}
\input{sec/2_method}
\input{sec/3_experiments}
\input{sec/4_conclusion}


\section*{Impact Statement}
This paper presents work whose goal is to advance understanding of low-level visual perception in vision-language models. DistortBench provides a diagnostic benchmark that can help identify and address perceptual gaps in current VLMs. Potential societal consequences include improved quality monitoring, content moderation, and image restoration systems. We do not foresee specific harmful applications; the primary impact is scientific and infrastructural.

\bibliography{main}
\bibliographystyle{icml2026}

\newpage
\appendix
\onecolumn
\input{sec/5_suppl}

\end{document}

%% file: sec/0_abstract.tex
\begin{abstract}
Vision-language models (VLMs) are increasingly used in settings where sensitivity to low-level image degradations matters, including content moderation, image restoration, and quality monitoring. Yet their ability to recognize distortion type and severity remains poorly understood. We present \textbf{DistortBench}, a diagnostic benchmark for no-reference distortion perception in VLMs. DistortBench contains 13{,}500 four-choice questions covering 27 distortion types, six perceptual categories, and five severity levels: 25 distortions inherit KADID-10k calibrations, while two added rotation distortions use monotonic angle-based levels. We evaluate 18 VLMs, including 17 open-weight models from five families and one proprietary model. Despite strong performance on high-level vision-language tasks, the best model reaches only 61.9\% accuracy, just below the human majority-vote baseline of 65.7\% (average individual: 60.2\%), indicating that low-level perceptual understanding remains a major weakness of current VLMs. Our analysis further reveals weak and non-monotonic scaling with model size, performance drops in most base--thinking pairs, and distinct severity-response patterns across model families. We hope DistortBench will serve as a useful benchmark for measuring and improving low-level visual perception in VLMs.
\end{abstract}

%% file: sec/1_intro.tex
\section{Introduction}
\label{sec:intro}

Image quality assessment (IQA) is a longstanding problem in computer vision with applications spanning content delivery~\cite{bosse2018deep}, image restoration~\cite{zhang2021designing}, and media forensics~\cite{verdoliva2020media}.
Traditional IQA methods focus on predicting a scalar quality score, either by comparing against a reference image (full-reference IQA~\cite{wang2004image,zhang2018unreasonable}) or directly from the distorted image alone (no-reference IQA~\cite{mittal2012no,ke2021musiq}).
However, a finer-grained understanding, \emph{what} distortion is present and \emph{how severe} it is, remains critical for downstream applications such as adaptive compression, automated image restoration pipelines, and quality-aware content moderation.

The rise of vision language models (VLMs)~\cite{liu2023visual,bai2025qwen25vl,team2025gemma3,chen2024internvl25} has introduced a new paradigm for visual understanding.
These models combine visual encoders with large language models to reason about images through natural language, achieving remarkable performance on tasks from visual question answering to document understanding.
Yet their capabilities for \emph{low-level} visual perception, the ability to detect and characterize fine-grained image degradations, remain underexplored.
Existing VLM benchmarks~\cite{wu2024qbench,zhang2024benchmark} have begun probing quality perception, but typically focus on coarse quality ratings or pairwise comparisons rather than fine-grained distortion identification.

In this work, we introduce \textbf{DistortBench}, a benchmark designed to evaluate VLMs on the joint task of distortion type identification and severity level estimation in a \emph{no-reference} setting.
Given only the distorted image, without access to the original, the model must select the correct combination of distortion type (from 27 possibilities across six categories) and severity level (from five levels) in a four-choice multiple-choice format.
This no-reference formulation is a deliberate design choice: it tests whether VLMs have internalized sufficient knowledge of natural image statistics and distortion signatures to identify degradations from appearance alone, mirroring real-world scenarios where pristine references are unavailable.
The task simultaneously tests the model's ability to recognize the \emph{type} of degradation and gauge its \emph{intensity}, a significantly more challenging task than binary quality classification or scalar score prediction.

We evaluate 17 open-weight VLMs spanning five model families (Qwen3.5, Gemma~3, InternVL3.5, Kimi-VL, and Qwen2-VL) with parameter counts ranging from 3B to 38B, including base and thinking variants for Qwen3.5 and Kimi-VL, plus one proprietary model: GPT-5.4~\cite{openai2025gpt5}.
Our analysis yields several insights:

\begin{itemize}[leftmargin=*,nosep]
    \item \textbf{The task is far from solved.} The best-performing model, Qwen3.5~27B (base), achieves 61.9\% accuracy, well above the 25\% chance baseline but leaving substantial room for improvement, revealing a significant gap in VLM low-level visual understanding.

    \item \textbf{Scaling is non-monotonic.} Larger models do not consistently outperform smaller ones. Within the Qwen3.5 base family, the 27B dense model strongly outperforms all other sizes including the 35B MoE variant. InternVL3.5~14B underperforms the 8B variant. The compact Kimi-VL~A3B (3B parameters) matches InternVL3.5~38B despite having $12.7\times$ fewer parameters.

    \item \textbf{Distortion difficulty spans a wide spectrum.} Mean accuracy across models ranges from 67\% on Gaussian blur and 66\% on impulse noise down to below 27\% on denoising artifacts, with rotation distortions exhibiting the highest cross-model variance ($\sim$19\% std), indicating architecture-dependent sensitivity.

    \item \textbf{Base models exhibit a U-shaped severity curve.} Non-thinking models perform well at both extremes (barely perceptible level~1 and clearly visible level~5) but dip at mid-range severity levels~2--3. Thinking variants show a more monotonic increase, indicating a different severity-response profile rather than straightforwardly eliminating this non-linearity.

    \item \textbf{Chain-of-thought reasoning usually hurts on perceptual tasks.} Thinking variants underperform their base counterparts in four of five paired comparisons, including across three of four Qwen3.5 sizes (e.g., Qwen3.5~27B Think.\ 58.2\% vs.\ base 61.9\%) and for Kimi-VL (44.0\% vs.\ 52.4\%), suggesting that deliberative reasoning often introduces noise on tasks that rely on direct pattern matching.

    \item \textbf{GPT-5.4 shows the steepest severity climb of any model.} GPT-5.4 rises monotonically from 38.4\% at level~1 to 72.9\% at level~5, a 34.5-point span that brings it from well below average at the lowest severity to among the top performers at the highest, a pattern distinct from all open-weight families and discussed further in \cref{sec:severity}.
\end{itemize}

\subsection{Related Work}
\label{sec:related}

\paragraph{Image quality assessment:}
IQA methods broadly fall into two paradigms.
Full-reference (FR) methods like SSIM~\cite{wang2004image} and LPIPS~\cite{zhang2018unreasonable} measure perceptual similarity between reference and distorted images, while no-reference (NR) methods~\cite{mittal2012no,ke2021musiq,su2020blindly} predict quality scores from distorted images alone.
IQA datasets such as LIVE~\cite{sheikh2006statistical}, CSIQ~\cite{larson2010most}, TID2013~\cite{ponomarenko2015image}, and KADID-10K~\cite{lin2019kadid} provide human-annotated quality scores for synthetically distorted images.
DistortBench draws on the distortion taxonomy of KADID-10K but reformulates the task from score prediction to distortion identification, adopting the more challenging NR setting where models must identify distortions without seeing the pristine original.

\paragraph{VLMs for quality assessment:}
Recent work has explored VLMs for image quality tasks.
Q-Bench~\cite{wu2024qbench} evaluates quality perception through low-level visual question answering.
Q-Align~\cite{wu2023qalign} fine-tunes VLMs for quality scoring.
DepictQA~\cite{you2024depicting} uses VLMs for multi-modal quality description.
These efforts primarily target overall quality perception; DistortBench instead focuses on the more diagnostic task of identifying specific distortion types and their severity levels.

\paragraph{VLM benchmarks:}
General-purpose VLM benchmarks~\cite{liu2023mmbench,yue2024mmmu,lu2022learn} assess high-level visual reasoning but do not systematically probe low-level perceptual capabilities.
Our benchmark complements these by providing a controlled evaluation of fine-grained visual sensitivity to image degradations.

%% file: sec/2_method.tex
\section{DistortBench}
\label{sec:method}

DistortBench evaluates vision language models on joint distortion type and severity level identification through a controlled multiple-choice assessment.
This section describes the distortion taxonomy (\cref{sec:distortions}), the dataset construction pipeline (\cref{sec:dataset}), and the evaluation protocol (\cref{sec:protocol}).

\subsection{Distortion Taxonomy}
\label{sec:distortions}

We define 27 distortion types organized into six perceptual categories, following the taxonomy established by KADID-10K~\cite{lin2019kadid}.
\cref{tab:distortions} summarizes the complete taxonomy.

\begin{table}[t]
\centering
\caption{\textbf{Distortion taxonomy.} 27 distortion types organized into six perceptual categories, each applied at five severity levels from very mild (level~1) to very strong (level~5).}
\label{tab:distortions}
\small
\setlength{\tabcolsep}{4pt}
\begin{tabular}{@{}llc@{}}
\toprule
\textbf{Category} & \textbf{Distortion Types} & \textbf{Count} \\
\midrule
Blur & Gaussian, Lens, Motion & 3 \\
Color & \makecell[l]{Diffusion, Shift, Quantization,\\ Saturation (over/de)} & 5 \\
Compression & JPEG, JPEG2000 & 2 \\
Noise & \makecell[l]{White, White (color), Impulse,\\ Multiplicative, Denoise} & 5 \\
Luminance & Brighten, Darken, Mean shift, Contrast & 4 \\
Spatial & \makecell[l]{Jitter, Non-eccentricity, Pixelate,\\ Quantization, Color block, Sharpen,\\ Rotation (CW/CCW)} & 8 \\
\midrule
\multicolumn{2}{@{}l}{\textbf{Total}} & \textbf{27} \\
\bottomrule
\end{tabular}
\end{table}

Each distortion type is applied at five severity levels, with level-specific parameters calibrated to produce perceptually distinguishable degradation steps, from barely perceptible (level~1) to clearly visible (level~5).
The distortion functions are implemented using standard image processing operations (OpenCV, SciPy) to ensure reproducibility; full parameter tables are provided in \cref{app:params}.

\paragraph{Severity calibration:}
Severity labels are operationally defined by the generation parameters used to produce each distorted image, making the ground truth unambiguous.
For the 25 distortion types inherited from KADID-10k~\cite{lin2019kadid}, the five parameter levels were calibrated to produce linearly spaced perceptual degradation steps spanning the full quality range, then validated via crowdsourced Degradation Mean Opinion Scores (DMOS) from 30 raters, which correlated at SROCC~$= 0.923$ with lab-based Mean Opinion Scores (MOS) from TID2013~\cite{ponomarenko2015image}.
As an additional sanity check, we aggregated the raw KADID crowd-rating records for the 10{,}125 inherited distortions and found that mean human quality scores decrease monotonically with severity overall (4.08$\to$3.52$\to$3.06$\to$2.50$\to$2.01 from levels~1--5), while 22 of the 25 inherited distortion types exhibit monotonic non-increasing mean scores across levels (details in \cref{app:kadid_human}).
The two rotation distortions (clockwise and anti-clockwise) are author-contributed additions not present in KADID-10k; their five levels correspond to $2^\circ, 5^\circ, 10^\circ, 20^\circ$, and $45^\circ$, a monotonically increasing sequence whose perceptual ordering is geometrically unambiguous, though not crowd-validated.
This does not constitute a direct human baseline for DistortBench's MCQ task, but it provides additional evidence that the inherited severity levels are aligned with human perceptual judgments.
We acknowledge that a model's internal severity scale may not align with this human-grounded calibration: a model that systematically identifies Level~3 distortions as Level~2 is not perceiving incorrectly in an absolute sense, but is miscalibrated relative to human perceptual consensus.
DistortBench makes this miscalibration directly measurable through its same-type/different-level distractors (\cref{sec:dataset}), which isolate severity discrimination from distortion-type recognition and expose the direction and magnitude of any systematic bias in a model's severity scale.

\begin{figure*}[t]
    \centering
    \includegraphics[width=\textwidth,height=0.22\textheight,keepaspectratio]{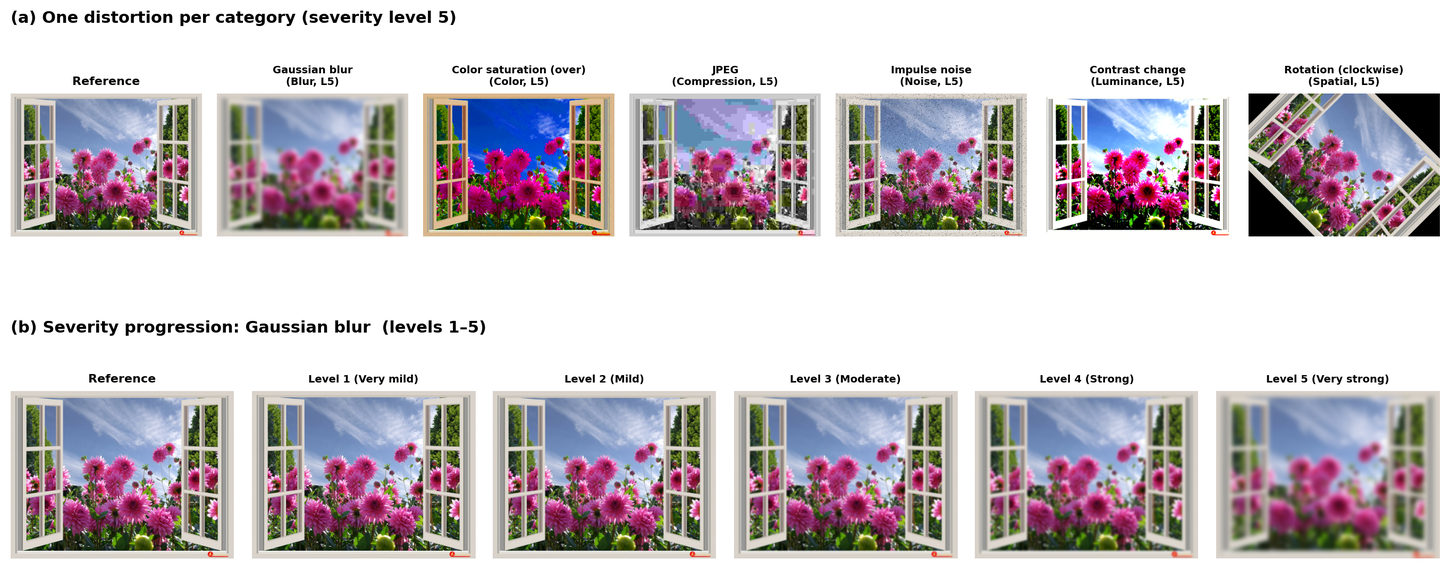}
    \caption{\textbf{DistortBench dataset examples.} (a)~A reference image and one representative distortion per category at severity level~5 (very strong), showing the perceptual diversity across the six distortion categories. (b)~Severity progression for Gaussian blur from level~1 (very mild) to level~5 (very strong). Models receive only the distorted image (no reference) and must jointly identify the distortion type and severity level.}
    \label{fig:dataset_examples}
\end{figure*}

\cref{fig:dataset_examples} illustrates the benchmark visually.
Panel~(a) shows a reference image alongside one representative distortion from each of the six categories at the strongest severity (level~5), highlighting the perceptual diversity across categories.
Panel~(b) shows a single distortion type (Gaussian blur) across all five severity levels, illustrating the progression from barely perceptible to clearly visible degradation that models must discriminate.

The six categories capture complementary aspects of image degradation:
\textbf{Blur} distortions reduce spatial frequency content through Gaussian, lens, and motion models;
\textbf{Color} distortions alter chromatic properties through diffusion, shifting, quantization, and saturation changes;
\textbf{Compression} distortions introduce codec-specific artifacts from JPEG and JPEG2000;
\textbf{Noise} distortions add various noise patterns (Gaussian, impulse, multiplicative) and include denoising artifacts;
\textbf{Luminance} distortions modify brightness and contrast;
and \textbf{Spatial} distortions alter geometric structure through jitter, pixelation, rotation, and sharpening.

\subsection{Dataset Construction}
\label{sec:dataset}

\paragraph{Image generation:}
We apply all $27 \times 5 = 135$ distortion--level combinations to a set of 100 reference images, producing 13{,}500 distorted images.
Reference images are a 100-image subset randomly sampled from the KADIS-700k pristine image collection~\cite{lin2019kadid}, which comprises 140{,}000 high-quality photographs sourced from Pixabay\footnote{\url{https://pixabay.com}; distributed under the Pixabay License, which permits free use, modification, and redistribution for commercial and non-commercial purposes.}. The underlying collection spans diverse content types (natural scenes, objects, textures), helping reduce content-specific bias.

\paragraph{MCQ generation:}
For each distorted image, we generate a four-choice multiple-choice question asking the model to jointly identify the distortion type and severity level.
The correct answer combines the distortion name and severity level (e.g., ``Gaussian blur -- Level~3 (Moderate)'').
Three distractors are sampled using a stratified strategy to create plausible wrong answers:

\begin{enumerate}[leftmargin=*,nosep]
    \item \textbf{Same type, different level:} Tests severity discrimination (e.g., ``Gaussian blur -- Level~1 (Very mild)'').
    \item \textbf{Different type, same level:} Tests type discrimination (e.g., ``Lens blur -- Level~3 (Moderate)'').
    \item \textbf{Different type, different level:} A fully contrastive distractor (e.g., ``JPEG -- Level~5 (Very strong)'').
\end{enumerate}

\begin{figure}[t]
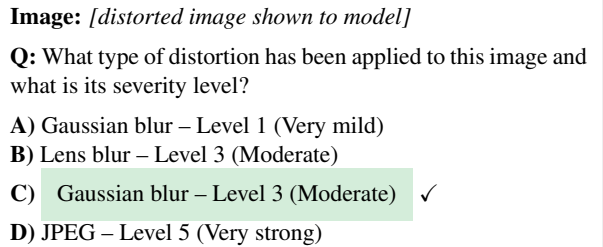

\centering
\small
\setlength{\fboxsep}{6pt}
\fbox{\parbox{0.92\linewidth}{
\textbf{Image:} \textit{[distorted image shown to model]}\\[4pt]
\textbf{Q:} What type of distortion has been applied to this image and what is its severity level?\\[4pt]
\textbf{A)} Gaussian blur -- Level 1 (Very mild)\\
\textbf{B)} Lens blur -- Level 3 (Moderate)\\
\textbf{C)} \colorbox{bestcolor}{Gaussian blur -- Level 3 (Moderate)}\;\checkmark\\
\textbf{D)} JPEG -- Level 5 (Very strong)
}}
\caption{\textbf{Example MCQ annotation.} The model sees only the distorted image and four choices. Distractors are stratified: same-type/different-level (A), different-type/same-level (B), and fully contrastive (D). The correct answer (C, highlighted) combines distortion type and severity.}
\label{fig:mcq_example}
\end{figure}

The four options are randomly shuffled across positions A--D to prevent position bias.
Question prompts are drawn from a pool of eight paraphrases to reduce sensitivity to specific phrasing.
\cref{fig:mcq_example} shows an example annotation.

\subsection{Evaluation Protocol}
\label{sec:protocol}

\paragraph{No-reference design:}
A central design decision in DistortBench is that models receive \emph{only the distorted image}, the original reference is never shown.
We adopt this no-reference (NR) setting for three reasons.
First, \emph{practical relevance}: in real-world deployment scenarios, content moderation, automated restoration pipelines, quality monitoring, the pristine original is typically unavailable, making NR identification the operationally meaningful task.
Second, \emph{diagnostic value}: providing the reference would reduce much of the task to visual differencing, testing the model's ability to spot pixel-level changes between two images rather than its understanding of what distortions \emph{look like}.
The NR setting instead probes whether VLMs have internalized sufficient knowledge of natural image statistics and distortion signatures to recognize degradations from appearance alone.
Third, \emph{alignment with human perception}: human observers routinely identify distortions (``this image looks blurry,'' ``there are compression artifacts'') without access to a reference, relying on learned priors about how natural images should appear.
DistortBench tests whether VLMs have acquired analogous perceptual priors.


\paragraph{Human baseline:}
To provide a human reference point, three annotators, each holding at least a master's degree and having over one year of experience in the imaging domain, independently answered the MCQ for a randomly selected subset of 256 distorted images under identical no-reference conditions. The majority-vote accuracy was \textbf{65.7\%} (95\% CI: [59.4, 71.5\%]), with a mean individual accuracy of 60.2\%. Inter-annotator agreement was moderate (Fleiss~$\kappa = 0.46$), and 9\% of samples failed to reach majority consensus; a complete per-distortion breakdown is provided in \cref{app:human_per_distortion}. A snapshot of the human annotation study is shown in \cref{fig:human_annotation_study}.


\paragraph{Model interface:}
All models receive the same input: a system prompt identifying the model as an ``expert image quality assessor,'' followed by the distorted image and the four-choice question.
Models are instructed to respond with only a JSON code block containing the answer letter (A/B/C/D), enabling automated parsing.
For thinking-capable models, we enable their native reasoning mode but keep the final answer format identical to the base models: these models may emit intermediate reasoning tokens before the final answer, after which the same extraction pipeline is applied to recover the terminal answer letter.
We use a three-stage answer extraction pipeline: (1)~JSON code block parsing, (2)~bare JSON parsing, and (3)~regex fallback, to maximize response coverage.
In practice, some thinking runs exhaust the generation budget before producing the terminal JSON answer, yielding a response that is valid text but unusable for scoring.

\paragraph{Metrics:}
We report three accuracy metrics.
\textbf{Joint accuracy} is the primary metric: the answered-only fraction of parseable responses for which the model selects the correct distortion type \emph{and} severity level.
\textbf{Type-balanced accuracy} macro-averages answered-only joint accuracy across the 27 distortion types, weighting each type equally regardless of sample count.
\textbf{Level-balanced accuracy} macro-averages answered-only joint accuracy across the five severity levels.
Because DistortBench is exactly balanced across distortion types and severity levels by construction, exact-match joint accuracy is the most interpretable primary metric; macro-averaged F1 is therefore omitted as it would provide little additional information.
We additionally provide per-distortion-type, per-category, and per-severity-level breakdowns to enable diagnostic analysis.
All accuracy figures include bootstrap 95\% confidence intervals (10{,}000 resamples).
Items where the model fails to produce a recoverable final answer (e.g., content-filter refusals, malformed outputs, or generations that exhaust the token budget before emitting the JSON answer) are excluded from the accuracy denominator; we report the \textbf{unparseable-response rate} (fraction of prompts with no recoverable answer over all 13{,}500 prompts) per model for transparency.

\paragraph{Infrastructure:}
Open-weight models are served locally using vLLM~\cite{kwon2023efficient} with bfloat16 precision.
Models with $>$14B parameters use tensor parallelism across two GPUs.
All models are accessed through an OpenAI-compatible API, ensuring a uniform evaluation interface.
Assessment runs use concurrent evaluation with configurable parallelism (default: 8 workers) and batched processing for memory efficiency.

%% file: sec/3_experiments.tex
\section{Experiments}
\label{sec:experiments}

\subsection{Models}
\label{sec:models}

We evaluate 17 open-weight vision language models from five families plus one proprietary model (GPT-5.4), spanning a range of architectures and parameter counts.
\cref{tab:models} lists all evaluated models.

\begin{table}[t]
\centering
\caption{\textbf{Evaluated models.} 17 open-weight VLMs from five families plus one proprietary model. Parameters indicate total model size; Qwen3.5~35B and Kimi-VL~A3B are mixture-of-experts (MoE) models with 3B active parameters. ``Think.'' denotes chain-of-thought (thinking) variants.}
\label{tab:models}
\small
\setlength{\tabcolsep}{3pt}
\begin{tabular}{@{}llrl@{}}
\toprule
\textbf{Family} & \textbf{Model} & \textbf{Params} & \textbf{Architecture} \\
\midrule
\multirow{8}{*}{Qwen3.5} & Qwen3.5 4B         & 4B  & Dense \\
                          & Qwen3.5 4B Think.  & 4B  & Dense + CoT \\
                          & Qwen3.5 9B         & 9B  & Dense \\
                          & Qwen3.5 9B Think.  & 9B  & Dense + CoT \\
                          & Qwen3.5 27B        & 27B & Dense \\
                          & Qwen3.5 27B Think. & 27B & Dense + CoT \\
                          & Qwen3.5 35B        & 35B & MoE (A3B) \\
                          & Qwen3.5 35B Think. & 35B & MoE (A3B) + CoT \\
\midrule
\multirow{3}{*}{Gemma 3} & Gemma3 4B  & 4B  & Dense \\
                          & Gemma3 12B & 12B & Dense \\
                          & Gemma3 27B & 27B & Dense \\
\midrule
\multirow{3}{*}{InternVL3.5} & InternVL3.5 8B  & 8B  & Dense \\
                              & InternVL3.5 14B & 14B & Dense \\
                              & InternVL3.5 38B & 38B & Dense \\
\midrule
\multirow{2}{*}{Kimi-VL} & Kimi-VL A3B           & 3B & MoE (A3B) \\
                          & Kimi-VL A3B Think.    & 3B & MoE (A3B) + CoT \\
\midrule
Qwen2-VL & Qwen2-VL 7B & 7B & Dense \\
\midrule
OpenAI & GPT-5.4 & --- & Proprietary \\
\bottomrule
\end{tabular}
\end{table}

The \textbf{Qwen3.5} family~\cite{bai2025qwen25vl} provides four size points, including a mixture-of-experts variant (35B total / 3B active parameters), each evaluated in both base and thinking (chain-of-thought) modes.
\textbf{Gemma~3}~\cite{team2025gemma3} contributes three base sizes from 4B to 27B.
\textbf{InternVL3.5}~\cite{chen2024internvl25} offers three sizes from 8B to 38B.
\textbf{Kimi-VL}~\cite{team2025kimi} provides both a standard and a thinking variant at 3B active parameters.
\textbf{Qwen2-VL}~\cite{wang2024qwen2} serves as a generational baseline.
We additionally evaluate one proprietary model: \textbf{GPT-5.4}~\cite{openai2025gpt5} accessed via Azure OpenAI, providing a reference point from the closed-source frontier.

\subsection{Main Results}
\label{sec:main_results}

\cref{tab:main_results} presents the overall performance of all 18 models (a visual comparison is provided in \cref{app:metrics}).
Several observations stand out.

\begin{table}[t]
\centering
\caption{\textbf{Main results.} Answered-only joint accuracy (Acc.) with bootstrap 95\% confidence intervals, type-balanced accuracy (T-Bal.), level-balanced accuracy (L-Bal.), and unparseable-response rate (Unp.) over the full 13{,}500-prompt evaluation. Models sorted by accuracy. Best result in \textbf{bold}, second-best \underline{underlined}. Random baseline is 25.0\%. $\dagger$~denotes proprietary model. $\ddagger$~MoE model (Qwen3.5~35B: total parameters listed, 3B active; Kimi-VL~A3B: active parameters listed). ``Think.'' denotes chain-of-thought variant. All accuracies in \%; unparseable-response rate in \%.}
\label{tab:main_results}
\footnotesize
\setlength{\tabcolsep}{2pt}
\resizebox{\columnwidth}{!}{%
\begin{tabular}{@{}lrccccc@{}}
\toprule
\textbf{Model} & \textbf{Params} & \textbf{Acc.} & \textbf{95\% CI} & \textbf{T-Bal.} & \textbf{L-Bal.} & \textbf{Unp.} \\
\midrule
Qwen3.5 27B        & 27B & \textbf{61.9} & \scriptsize{[61.1, 62.7]} & \textbf{61.9} & \textbf{61.9} & 0.0 \\
Qwen3.5 27B Think. & 27B & \underline{58.2} & \scriptsize{[57.4, 59.0]} & \underline{58.3} & \underline{58.2} & 1.3 \\
Qwen3.5 35B$^\ddagger$        & 35B & 56.8 & \scriptsize{[56.0, 57.6]} & 56.8 & 56.8 & 0.0 \\
Qwen3.5 4B         & 4B  & 56.1 & \scriptsize{[55.3, 57.0]} & 56.1 & 56.1 & 0.0 \\
Qwen3.5 9B Think.  & 9B  & 55.0 & \scriptsize{[54.2, 55.9]} & 55.0 & 55.0 & 0.8 \\
Qwen3.5 35B Think.$^\ddagger$ & 35B & 54.6 & \scriptsize{[53.8, 55.5]} & 54.7 & 54.6 & 1.3 \\
Qwen3.5 4B Think.  & 4B  & 53.1 & \scriptsize{[52.2, 53.9]} & 53.1 & 53.0 & 1.5 \\
InternVL3.5 38B    & 38B & 52.6 & \scriptsize{[51.8, 53.5]} & 52.6 & 52.6 & 0.0 \\
Kimi-VL A3B$^\ddagger$        & 3B  & 52.4 & \scriptsize{[51.6, 53.2]} & 52.4 & 52.4 & 0.0 \\
Qwen3.5 9B         & 9B  & 51.8 & \scriptsize{[51.0, 52.7]} & 51.8 & 51.8 & 0.0 \\
GPT-5.4$^\dagger$  & --- & 51.6 & \scriptsize{[50.7, 52.4]} & 51.6 & 51.6 & 0.1 \\
Gemma3 27B         & 27B & 47.5 & \scriptsize{[46.7, 48.3]} & 47.5 & 47.5 & 0.0 \\
Gemma3 12B         & 12B & 46.6 & \scriptsize{[45.8, 47.5]} & 46.6 & 46.6 & 0.0 \\
InternVL3.5 8B     & 8B  & 44.4 & \scriptsize{[43.6, 45.3]} & 44.4 & 44.4 & 0.7 \\
Kimi-VL A3B Think.$^\ddagger$ & 3B  & 44.0 & \scriptsize{[43.1, 44.8]} & 43.9 & 43.8 & 3.4 \\
InternVL3.5 14B    & 14B & 43.4 & \scriptsize{[42.6, 44.2]} & 43.4 & 43.4 & 0.0 \\
Qwen2-VL 7B        & 7B  & 36.7 & \scriptsize{[35.9, 37.5]} & 36.7 & 36.7 & 0.0 \\
Gemma3 4B          & 4B  & 33.7 & \scriptsize{[32.9, 34.5]} & 33.7 & 33.7 & 0.0 \\
\bottomrule
\end{tabular}%
}
\end{table}

\paragraph{Human comparison:}
The human majority-vote baseline (65.7\%, 95\% CI: [59.4, 71.5]) sits above the best model (61.9\%), with the human point estimate exceeding the best model's 95\% CI upper bound (62.7\%); the wide human CI reflects the modest study size (n=233 consensus samples), so the gap should be interpreted as indicative rather than conclusive.
However, human \emph{average} accuracy (60.2\%) is already within reach of the top model and exceeded by only Qwen3.5~27B base.
The moderate inter-annotator agreement ($\kappa = 0.46$) and 9\% no-consensus rate (23/256 samples) highlight that the task is genuinely difficult even for humans, particularly at mid-range severity levels where models also struggle most.

\paragraph{Overall difficulty:}
The best model achieves 61.9\% answered-only accuracy (95\% CI: [61.1, 62.7]), representing a 36.9 percentage point improvement over random chance (25\%) but leaving 38.1\% of answered items incorrect.
This confirms that joint distortion--severity identification is a challenging perceptual task for current VLMs.

\paragraph{Qwen3.5 family dominance:}
The Qwen3.5 family dominates the leaderboard: base and thinking variants together hold the top six positions among all 18 models, with three base models in the top four.
Even the smallest Qwen3.5 base variant (4B, 56.1\%) surpasses the proprietary model and all other open-weight families, suggesting that this model family has markedly stronger low-level visual perception capabilities.

\paragraph{Other notable findings:}
Kimi-VL~A3B (3B active parameters, 52.4\%) and InternVL3.5~38B (52.6\%) have overlapping confidence intervals ([51.6, 53.2] vs.\ [51.8, 53.5]), so the difference is not statistically significant despite the $12.7\times$ parameter gap, demonstrating that efficient MoE architectures can be highly competitive on perceptual tasks.
GPT-5.4 (51.6\%, CI: [50.7, 52.4]) places mid-table, below the top-3 open-weight models; as discussed in \cref{sec:severity}, it exhibits a distinctive monotonically ascending severity profile unlike most open-weight models.
Qwen2-VL~7B (36.7\%) lags far behind similarly-sized Qwen3.5 models, highlighting generational progress in low-level visual capabilities.
Thinking variants show non-trivial unparseable-response rates (up to 3.4\% for Kimi-VL Think.), meaning their answered-only accuracies are computed on a slightly easier effective subset; their true performance gap relative to base variants may therefore be even larger than the reported accuracy numbers suggest.
For the thinking-capable open models, this is often not a deliberate refusal: some runs consume the 8192-token generation budget on reasoning tokens and never emit the final JSON answer required for scoring.

\subsection{Scaling Analysis}
\label{sec:scaling}

\begin{figure}[t]
    \centering
    \includegraphics[width=\linewidth]{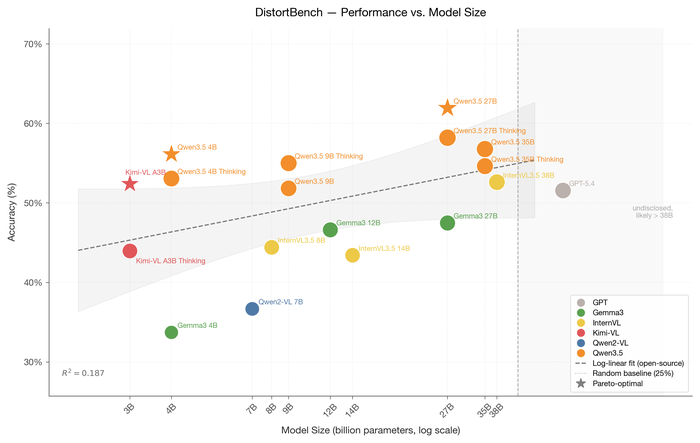}
    \caption{\textbf{Accuracy vs.\ model size.} Log-linear scaling plot across model families. Star markers indicate Pareto-optimal models. Scaling is broadly positive but weak, with non-monotonic behavior within Qwen3.5.}
    \label{fig:scaling}
\end{figure}

\cref{fig:scaling} shows accuracy as a function of model size (log scale) across families.
Within-family scaling behavior is inconsistent:

\begin{itemize}[leftmargin=*,nosep]
    \item \textbf{Qwen3.5 (base):} Accuracy is non-monotonic: 9B (51.8\%) $<$ 4B (56.1\%) $<$ 35B (56.8\%) $<$ 27B (61.9\%). The 27B model is a clear outlier above the trend; the 35B MoE model with only 3B active parameters underperforms the denser 27B despite having more total parameters, confirming that total parameter count is misleading for MoE models on perceptual tasks.
    \item \textbf{Qwen3.5 (thinking):} Also non-monotonic by parameter count: 4B (53.1\%) $\to$ 9B (55.0\%) $\to$ 27B (58.2\%) $\to$ 35B (54.6\%), with the 35B MoE again dropping below the 27B dense model.
    \item \textbf{Gemma~3:} Roughly monotonic scaling from 4B (33.7\%) to 12B (46.6\%) to 27B (47.5\%), though with diminishing returns, the 2.25$\times$ increase from 12B to 27B yields less than 1~pp gain, and the two models' CIs overlap ([45.8, 47.5] vs.\ [46.7, 48.3]).
    \item \textbf{InternVL3.5:} Non-monotonic with 8B (44.4\%) and 14B (43.4\%) having overlapping CIs ([43.6, 45.3] vs.\ [42.6, 44.2]), while 38B (52.6\%) provides a substantial and statistically significant jump.
\end{itemize}

The overall log-linear fit yields a low $R^2$, confirming that model size is a weak predictor of distortion identification performance.
Architecture, training data composition, and visual encoder quality appear to matter more than raw parameter count for low-level visual perception tasks.

Notably, despite the broad positive trend, the human majority-vote baseline (65.7\%) lies above all 18 evaluated models, and even the human individual average (60.2\%) is surpassed by only Qwen3.5~27B (61.9\%), reinforcing that raw parameter scaling alone is insufficient to close the perceptual gap.

\subsection{Per-Category Analysis}
\label{sec:per_category}

Performance varies substantially across the six distortion categories (full table and radar chart in \cref{app:per_category}).
Blur is the most accessible category, with all models above 44\%, while Luminance exhibits the widest range (19.4\% for Gemma3~4B to 59.2\% for Qwen3.5~27B).
Compression and Luminance are consistently the hardest categories, and Spatial distortions, despite including visually salient rotations and color blocks, show high cross-model variance due to confusable subtypes like jitter and non-eccentricity.

\subsection{Per-Distortion Difficulty}
\label{sec:difficulty}

\begin{figure}[t]
    \centering
    \includegraphics[width=\linewidth]{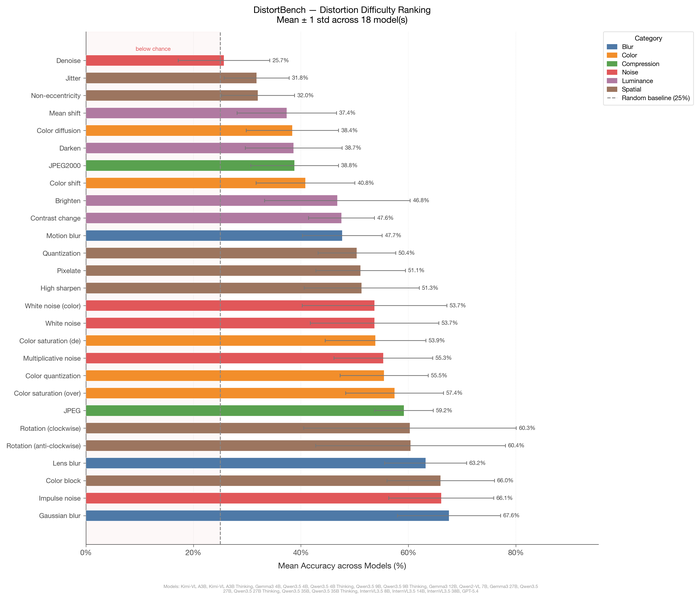}
    \caption{\textbf{Per-distortion difficulty ranking.} Mean accuracy $\pm$ standard deviation across all 18 models for each of the 27 distortion types, sorted by difficulty. The dashed line marks the 25\% random baseline.}
    \label{fig:difficulty}
\end{figure}

\cref{fig:difficulty} shows per-distortion accuracy averaged across all models.
The difficulty spectrum spans from 67.4\% (Gaussian blur) to 26.2\% (denoising artifacts), a 41-point range.
The easiest distortions (Gaussian blur, impulse noise, color block, lens blur) all produce visually distinctive artifacts; the hardest (denoise, jitter, non-eccentricity) produce subtle changes that closely resemble other distortion types.
Rotation distortions, despite ranking among the easiest on average, exhibit the highest cross-model variance ($\sim$19\% std for clockwise, $\sim$17\% for anti-clockwise); some models exceed 80\% while others fall below 40\%, indicating architecture-dependent geometric sensitivity.
Full difficulty analysis with consensus plots is provided in \cref{app:difficulty}.
The human difficulty ordering diverges markedly from the VLM ordering: humans achieve 100\% on rotations and Gaussian blur but find denoise far easier (83.3\%) than VLMs do (26.2\%); the full per-distortion human breakdown is in \cref{app:human_per_distortion}.

\subsection{Severity Level Analysis}
\label{sec:severity}

\begin{figure}[t]
    \centering
    \includegraphics[width=\linewidth]{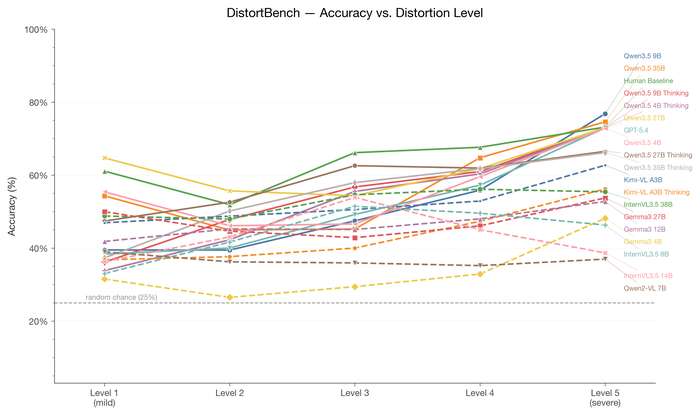}
    \caption{\textbf{Accuracy vs.\ distortion severity level.} Base models show a U-shaped curve (high at levels~1 and~5, dip at levels~2--3); thinking variants rise monotonically. GPT-5.4 shows the steepest monotonic climb (38.4\%$\to$72.9\%).}
    \label{fig:accuracy_level}
\end{figure}

\cref{fig:accuracy_level} shows accuracy as a function of severity level across all models.
A striking split emerges between base and thinking variants.
\textbf{Base models} exhibit a U-shaped curve: they perform relatively well at level~1 and level~5 but dip at the mid-range levels~2--3.
For Qwen3.5~27B (base), accuracy is 64.7\% at level~1, drops to 55.7\% at level~2 and 54.4\% at level~3, then recovers to 61.8\% at level~4 and 73.0\% at level~5.
Qwen3.5~4B (base) shows the same U-shape: 55.4\% at level~1, dipping to 46.1\% at level~2 before climbing to 72.9\% at level~5.
\textbf{Thinking variants}, in contrast, show a roughly monotonic increase from low to high severity, with Qwen3.5~27B (Think.) rising from 47.5\% at level~1 to 66.6\% at level~5.
GPT-5.4 follows a monotonic pattern similar to thinking variants (38.4\% $\to$ 72.9\%), showing the steepest climb of any model, a 34.5-point span that brings it from well below average at level~1 to among the top performers at level~5.
One possible explanation for the U-shaped pattern in base models is that two distinct competencies are at play: recognizing the \emph{signature} of a distortion at its most subtle extreme, and recognizing its \emph{full expression} at strong severity, with mid-range severity posing a harder discrimination problem.
However, this hypothesis requires further investigation to rule out alternative explanations such as distractor difficulty varying by level.
Notably, human annotators show an analogous severity-response pattern: majority-vote accuracy on consensus samples is 61.9\%, 53.8\%, 66.7\%, 63.8\%, and 77.2\% at levels~1--5, with the same level~2 dip, suggesting this non-monotonicity is an intrinsic property of the task rather than a VLM-specific artifact; humans surpass all models at level~5 (77.2\% vs.\ 73.0\% for Qwen3.5~27B) while several base models approach or exceed the human majority-vote accuracy at level~1.

\subsection{Error and Qualitative Analysis}
\label{sec:error}

\begin{figure}[t]
    \centering
    \includegraphics[width=\linewidth]{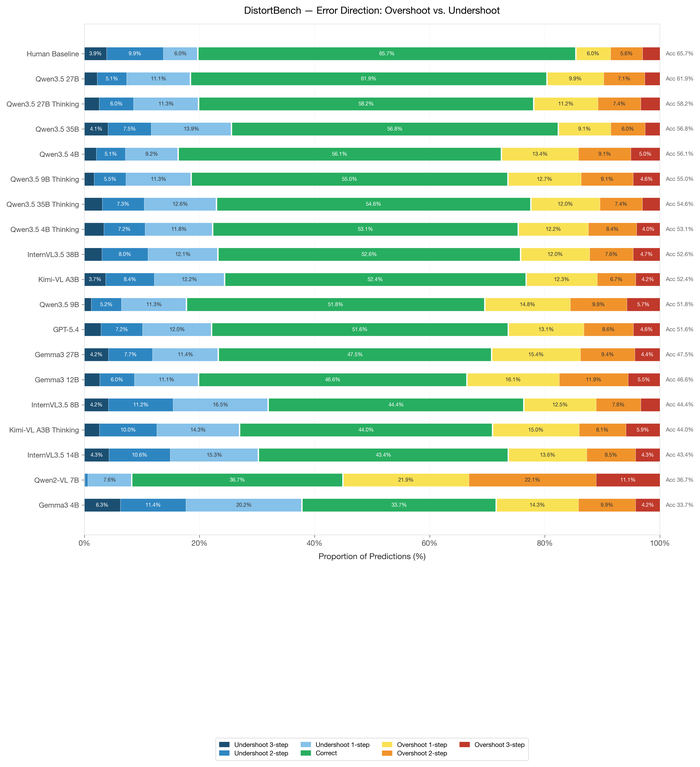}
    \caption{\textbf{Error direction analysis.} Stacked bars decomposing predictions into correct (green, center) and severity errors by step size: undershoots (left, blues) vs.\ overshoots (right, yellows/reds). Models sorted by accuracy.}
    \label{fig:error_direction}
\end{figure}

Most models exhibit balanced confusion patterns across the four answer choices (\cref{app:error}); Qwen2-VL~7B is the notable exception with extreme C/D bias.
\cref{fig:error_direction} decomposes errors by direction: severity undershoots (predicting a milder level than ground truth) consistently outweigh overshoots across all models, with the asymmetry strongest for weaker models.
Qualitative examples (\cref{app:qualitative}) confirm that easy distortions (rotation, Gaussian blur) produce unmistakable visual signatures even for mid-tier models, while hard distortions (non-eccentricity, jitter) defeat even the best models. They also show that, although some stronger base models remain relatively accurate at level~1 in aggregate, these very mild examples still frequently induce same-type/wrong-level confusions.
Human annotators exhibit the same error signature: individual type-only accuracy is 83.6\% on average, while level-only accuracy is 69.9\%, confirming that severity discrimination, not distortion-type recognition, is the dominant failure mode for both humans and models.

\subsection{Effect of Chain-of-Thought Reasoning}
\label{sec:finetuning}

Five base--thinking pairs (four Qwen3.5 sizes + Kimi-VL) enable a systematic comparison.
Four of five favor the base model, with degradations of $-$3.7~pp (Qwen3.5~27B), $-$3.0~pp (4B), $-$2.2~pp (35B), and $-$8.4~pp (Kimi-VL); only Qwen3.5~9B bucks the trend ($+$3.2~pp).
Thinking variants also exhibit higher unparseable-response rates (0.8--3.4\% vs.\ 0--0.7\% for all base models), meaning their answered-only accuracies are computed on a slightly easier effective test set after excluding samples with no recoverable final answer; the true accuracy gap may therefore be wider than reported.
This contrasts sharply with findings on high-level reasoning benchmarks~\cite{wei2022chain} and suggests that distortion identification relies on direct perceptual pattern matching rather than deliberative step-by-step reasoning; explicit reasoning chains introduce noise on a task where the answer is ``seen'' rather than ``derived.''

%% file: sec/4_conclusion.tex
\section{Conclusion}
\label{sec:conclusion}

We presented DistortBench, a benchmark for evaluating vision language models on joint distortion type and severity level identification.
Our evaluation of 17 open-weight VLMs spanning five model families, including base and thinking variants for Qwen3.5 and Kimi-VL and one proprietary model (GPT-5.4) reveals substantial room for improvement: the best model, Qwen3.5~27B (base), achieves 61.9\% joint accuracy on a four-choice task (25\% random baseline), with accuracy dropping below 35\% on the hardest distortion types.

Our analysis yields several actionable insights for the VLM community:
(1)~low-level visual perception does not scale straightforwardly with model size, suggesting that architectural innovations or targeted training data may be more impactful than simple parameter scaling;
(2)~the dramatic difficulty spectrum across distortion types (26\% to 67\%) points to specific perceptual gaps that could guide training data curation;
(3)~base models exhibit a U-shaped severity curve, where they perform well at both extremes but struggle at mid-range severity. Thinking variants show a more monotonic increase, revealing a qualitative difference in how reasoning modes handle perceptual degradations;
(4)~chain-of-thought reasoning usually hurts performance on perceptual tasks, with degradations in four of five base--thinking comparisons across the Qwen3.5 and Kimi-VL families, motivating research into alternative inference strategies for low-level vision;
and (5)~GPT-5.4 shows the steepest severity climb of any model (38.4\%$\to$72.9\%), a monotonically ascending profile qualitatively distinct from all open-weight families and warranting investigation into the underlying perceptual mechanisms.
We release DistortBench to support this effort.
Limitations and future directions are discussed in \cref{sec:limitations}.


%% file: sec/5_suppl.tex
\section*{Supplementary Material}
\setcounter{figure}{0}
\renewcommand{\thefigure}{A\arabic{figure}}
\setcounter{table}{0}
\renewcommand{\thetable}{A\arabic{table}}
\setlength{\textfloatsep}{12pt plus 2pt minus 4pt}
\setlength{\floatsep}{10pt plus 2pt minus 2pt}
\setlength{\intextsep}{10pt plus 2pt minus 2pt}
\newcommand{\appsectionspace}{\Needspace{0.45\textheight}}

This appendix provides extended quantitative breakdowns, qualitative examples, and distortion-generation details that support the paper.

\section{Limitations}
\label{sec:limitations}

DistortBench applies distortions synthetically to clean images; findings may not fully transfer to real-world corruptions from sensors, encoders, or transmission artifacts, and compound distortions are not covered.
Proprietary model coverage is limited to a single model (GPT-5.4); broader closed-source evaluation remains future work.
The human baseline rests on three annotators and 256 images, giving wide confidence intervals that weaken the human--model gap claim.
Finally, models trained on KADID or KADIS-derived data may have encountered benchmark images during training, and contamination effects are unquantified.
These limitations point to concrete future directions: extending DistortBench to real-world and compound distortions, scaling the human annotation study to establish a stronger ceiling, broadening proprietary model coverage, and leveraging the per-distortion difficulty signal to guide targeted VLM pre-training for low-level perception.

\appsectionspace
\section{Per-Category Analysis}
\label{app:per_category}

\begin{figure}[H]
    \centering
    \includegraphics[width=\linewidth]{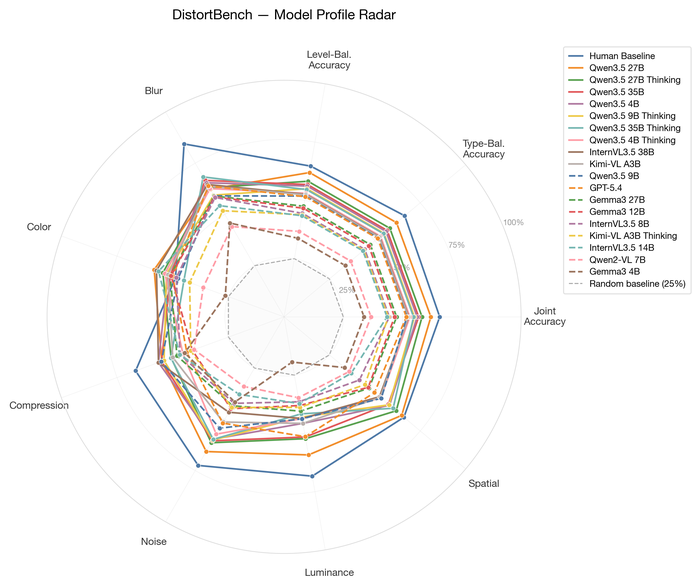}
    \caption{\textbf{Radar chart.} 9-axis profile combining three overall metrics (joint accuracy, type-balanced accuracy, level-balanced accuracy) and six per-category accuracies. The dashed circle marks the 25\% random baseline.}
    \label{fig:radar}
\end{figure}

\cref{fig:radar} shows 9-axis profiles combining overall metrics and per-category accuracy.
Blur is the strongest category overall, while Spatial distortions remain comparatively strong for some models due to easy subtypes such as rotations and color blocks; compression and luminance distortions are consistently challenging across all models.
The Qwen3.5 family shows the most expanded profiles, while Gemma3~4B and Qwen2-VL~7B (inner polygons) are clearly separated from the rest, particularly on color and luminance categories.

\cref{tab:per_category} provides the full per-category accuracy breakdown for all 18 models.
Blur is the most consistently accessible category (all models above 44\%), while Luminance exhibits the widest range (19.4\% for Gemma3~4B to 59.2\% for Qwen3.5~27B).

\begin{table}[H]
\centering
\caption{\textbf{Per-category accuracy (\%).} Accuracy for each of the six distortion categories across all 18 models, sorted by overall accuracy. Best per column in \textbf{bold}.}
\label{tab:per_category}
\small
\setlength{\tabcolsep}{2pt}
\resizebox{\linewidth}{!}{%
\begin{tabular}{@{}lcccccc@{}}
\toprule
\textbf{Model} & \textbf{Blur} & \textbf{Color} & \textbf{Comp.} & \textbf{Noise} & \textbf{Lum.} & \textbf{Spat.} \\
\midrule
Qwen3.5 27B          & 62.3 & \textbf{58.3} & 54.9 & \textbf{65.7} & \textbf{59.2} & \textbf{64.8} \\
Qwen3.5 27B Think.   & 63.0 & 54.4 & 50.7 & 61.4 & 52.3 & 61.7 \\
Qwen3.5 35B          & 66.6 & 51.1 & 55.0 & 60.5 & 51.5 & 57.5 \\
Qwen3.5 4B           & 65.5 & 52.1 & \textbf{56.4} & 59.7 & 45.7 & 58.1 \\
Qwen3.5 9B Think.    & 59.4 & 52.8 & 53.9 & 60.1 & 43.0 & 57.8 \\
Qwen3.5 35B Think.   & \textbf{68.3} & 45.0 & 50.8 & 59.8 & 41.6 & 60.1 \\
Qwen3.5 4B Think.    & 62.6 & 53.5 & 45.4 & 57.2 & 43.6 & 53.3 \\
InternVL3.5 38B      & 64.6 & 57.3 & 56.4 & 46.6 & 43.6 & 52.5 \\
Kimi-VL A3B          & 63.7 & 52.1 & 50.0 & 51.3 & 45.7 & 53.0 \\
Qwen3.5 9B           & 59.1 & 47.6 & 55.1 & 54.3 & 43.8 & 53.5 \\
GPT-5.4              & 63.9 & 50.0 & 43.7 & 51.8 & 51.4 & 49.8 \\
Gemma3 27B           & 59.0 & 51.1 & 48.1 & 43.4 & 40.5 & 46.8 \\
Gemma3 12B           & 58.6 & 50.8 & 41.2 & 44.8 & 38.0 & 46.3 \\
InternVL3.5 8B       & 58.3 & 48.4 & 46.7 & 42.2 & 36.4 & 41.5 \\
Kimi-VL Think.       & 51.8 & 42.4 & 41.9 & 44.2 & 38.8 & 44.5 \\
InternVL3.5 14B      & 54.3 & 56.4 & 46.8 & 37.8 & 37.2 & 37.0 \\
Qwen2-VL 7B          & 44.1 & 36.3 & 40.4 & 34.0 & 34.7 & 35.9 \\
Gemma3 4B            & 45.8 & 26.4 & 44.6 & 41.6 & 19.4 & 33.4 \\
\bottomrule
\end{tabular}
}
\end{table}

\appsectionspace
\section{Per-Distortion Difficulty Analysis}
\label{app:difficulty}

The per-distortion difficulty ranking (shown in the paper, \cref{fig:difficulty}) spans a remarkably wide spectrum.
Here we provide the detailed breakdown:

\paragraph{Easiest distortions ($>$58\% mean accuracy).}
Gaussian blur (67.4\%), impulse noise and color block (both 66.0\%), lens blur (63.7\%), rotation (clockwise: 60.0\%, anti-clockwise: 59.9\%), and JPEG (58.7\%) are the most reliably identified.
These distortions produce visually distinctive artifacts: impulse noise creates scattered bright/dark pixels, blur distortions reduce high-frequency detail in characteristic ways, and color blocks introduce regular patch patterns.

\paragraph{Hardest distortions ($<$35\% mean accuracy).}
Denoise (26.2\%), jitter (32.1\%), and non-eccentricity (33.2\%) are consistently difficult.
Denoise is the hardest distortion overall, with mean accuracy only slightly above the 25\% random baseline and some models falling below it; its over-smoothing artifacts closely resemble mild blur, making it nearly indistinguishable.
Non-eccentricity produces subtle barrel/pincushion warping easily confused with other spatial transforms.
Jitter introduces micro-displacements that may fall below the perceptual resolution of current VLM visual encoders.

\begin{figure}[H]
    \centering
    \includegraphics[width=\linewidth]{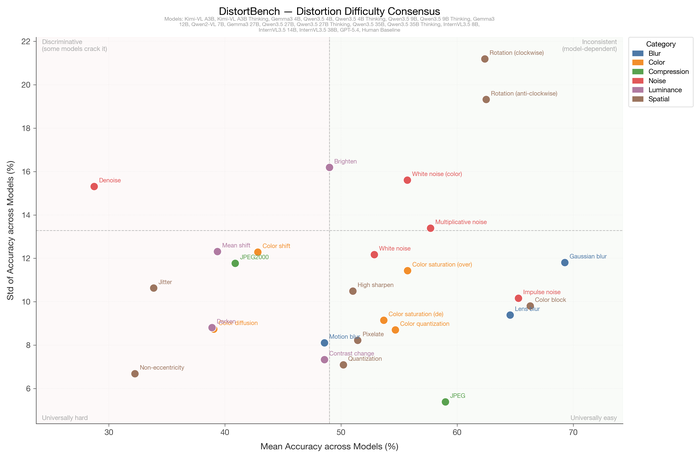}
    \caption{\textbf{Difficulty consensus.} Mean accuracy vs.\ cross-model standard deviation for each distortion type. High-consensus distortions (low std) cluster at accuracy extremes.}
    \label{fig:consensus}
\end{figure}

\paragraph{Model consensus.}
\cref{fig:consensus} reveals that cross-model agreement is highest at the difficulty extremes: JPEG, jitter, and contrast change cluster at low standard deviation, indicating they are ``universally easy'' or ``universally hard'' regardless of model.
Rotation distortions, despite ranking among the easiest on average, exhibit the \emph{highest} cross-model variance ($\sim$19\% std for CW), placing them in the ``inconsistent / model-dependent'' quadrant; some models achieve $>$80\% while others struggle below 40\%.
Brighten is another high-variance outlier, suggesting that luminance sensitivity varies dramatically across architectures.

\clearpage
\section{Human Baseline Per-Distortion Results}
\label{app:human_per_distortion}

\cref{tab:human_per_distortion} reports individual human accuracy for each of the 27 distortion types, sorted by accuracy.
\cref{tab:human_per_severity} provides a complementary breakdown by severity level.
The human difficulty ordering differs markedly from the VLM ordering.
Humans achieve perfect accuracy (100\%) on Gaussian blur, multiplicative noise, and both rotation distortions, distortions with unambiguous visual signatures.
Notably, humans find \emph{denoise} much easier (83.3\%) than models do (mean 26.2\%), suggesting that the over-smoothing artifacts are perceptually recognisable to humans but fall outside the pattern-matching repertoire of current VLMs.
Conversely, humans struggle most on color saturation (over) (25.0\%), non-eccentricity (36.4\%), and white noise (37.5\%), categories where models are not uniformly the weakest.
The broad agreement between human and model difficulty at the extremes (rotations easy; luminance/color subtypes hard) alongside the denoise divergence highlights specific perceptual gaps that targeted VLM training could address.

\begin{table}[H]
\centering
\caption{\textbf{Human per-distortion accuracy.} Majority-vote accuracy for each of the 27 distortion types over the 256-image human annotation study (3 annotators). Sorted by accuracy descending. Fractions show consensus samples per type (samples where $\geq$2/3 annotators agreed); note sample counts are unequal due to random sampling.}
\label{tab:human_per_distortion}
\small
\setlength{\tabcolsep}{5pt}
\begin{tabular}{@{}lcc@{}}
\toprule
\textbf{Distortion Type} & \textbf{Correct / Consensus} & \textbf{Accuracy (\%)} \\
\midrule
Gaussian blur             &  7/7  & 100.0 \\
Multiplicative noise      &  5/5  & 100.0 \\
Rotation (clockwise)      &  9/9  & 100.0 \\
Rotation (anti-clockwise) &  7/7  & 100.0 \\
White noise (color)       & 11/12 &  91.7 \\
Brighten                  &  8/9  &  88.9 \\
Lens blur                 &  8/9  &  88.9 \\
Denoise                   &  5/6  &  83.3 \\
Color shift               & 11/14 &  78.6 \\
JPEG2000                  &  7/9  &  77.8 \\
Mean shift                &  3/4  &  75.0 \\
Jitter                    &  5/7  &  71.4 \\
Color block               &  5/7  &  71.4 \\
Contrast change           &  6/9  &  66.7 \\
Motion blur               &  9/14 &  64.3 \\
Pixelate                  &  4/7  &  57.1 \\
JPEG                      &  5/9  &  55.6 \\
Color diffusion           &  4/8  &  50.0 \\
Impulse noise             &  3/6  &  50.0 \\
Color saturation (de)     &  4/8  &  50.0 \\
Quantization              &  6/13 &  46.2 \\
High sharpen              &  5/11 &  45.5 \\
Darken                    &  3/7  &  42.9 \\
Color quantization        &  2/5  &  40.0 \\
White noise               &  6/16 &  37.5 \\
Non-eccentricity          &  4/11 &  36.4 \\
Color saturation (over)   &  1/4  &  25.0 \\
\midrule
\textbf{Overall (majority vote)} & \textbf{153/233} & \textbf{65.7} \\
\bottomrule
\end{tabular}
\end{table}

\begin{table}[H]
\centering
\caption{\textbf{Human per-severity majority-vote accuracy.} Majority-vote accuracy on consensus samples (i.e., samples where at least 2 of 3 annotators agreed on the same answer) for each severity level. No-consensus counts indicate samples excluded from the denominator due to three-way disagreement.}
\label{tab:human_per_severity}
\small
\setlength{\tabcolsep}{4pt}
\begin{tabular}{@{}lcccc@{}}
\toprule
\textbf{Severity Level} & \textbf{Correct / Consensus} & \textbf{Accuracy (\%)} & \textbf{No-Consensus} \\
\midrule
Level 1 (Very mild)   & 26/42 & 61.9 & 8 \\
Level 2 (Mild)        & 21/39 & 53.8 & 9 \\
Level 3 (Moderate)    & 32/48 & 66.7 & 1 \\
Level 4 (Strong)      & 30/47 & 63.8 & 4 \\
Level 5 (Very strong) & 44/57 & 77.2 & 1 \\
\midrule
\textbf{Overall}      & \textbf{153/233} & \textbf{65.7} & \textbf{23} \\
\bottomrule
\end{tabular}
\end{table}

\clearpage
\section{Error Analysis}
\label{app:error}

\begin{figure}[H]
    \centering
    \includegraphics[width=\linewidth,height=0.68\textheight,keepaspectratio]{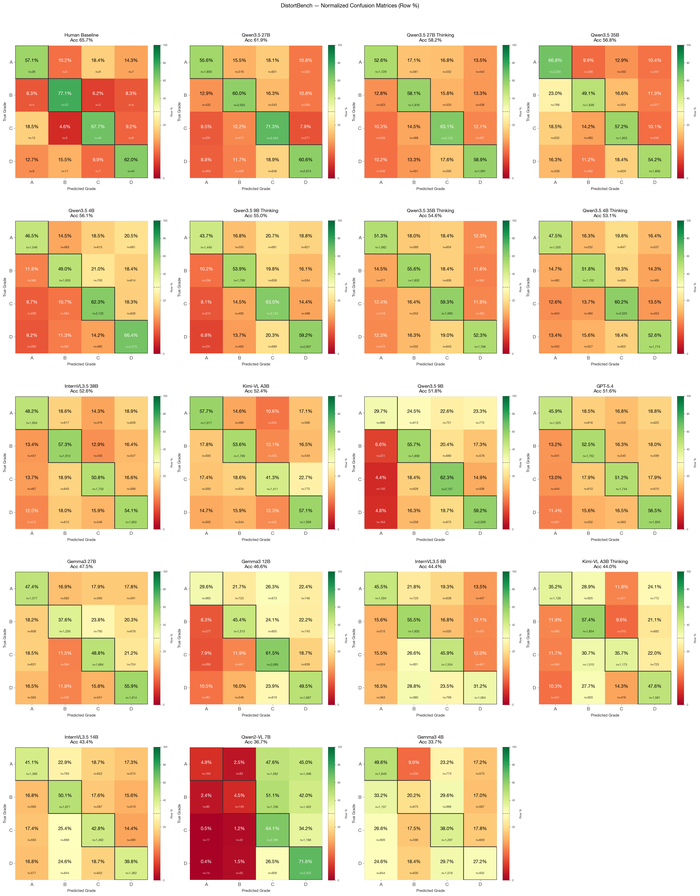}
    \caption{\textbf{Row-normalized position-bias confusion matrices} for all evaluated models. Since answer options are randomly shuffled, these matrices diagnose position bias rather than distortion-type confusions. Diagonal dominance indicates no position preference; off-diagonal mass reveals systematic position biases. Qwen2-VL~7B shows extreme C/D position bias.}
    \label{fig:confusion}
\end{figure}

\paragraph{Position-bias patterns.}
\cref{fig:confusion} shows row-normalized confusion matrices over the four answer positions (A--D).
Since answer options are randomly shuffled, these matrices diagnose \emph{position bias}, whether a model favors certain option positions regardless of content, rather than systematic distortion-type confusions.
Most models exhibit relatively balanced prediction patterns across positions, with errors distributed roughly uniformly across incorrect options.
The notable exception is Qwen2-VL~7B, which shows extreme bias toward positions C and D, suggesting the model has difficulty following the structured response format.

\clearpage
\section{Qualitative Analysis}
\label{app:qualitative}

\begin{figure}[H]
    \centering
    \includegraphics[width=\linewidth,height=0.82\textheight,keepaspectratio]{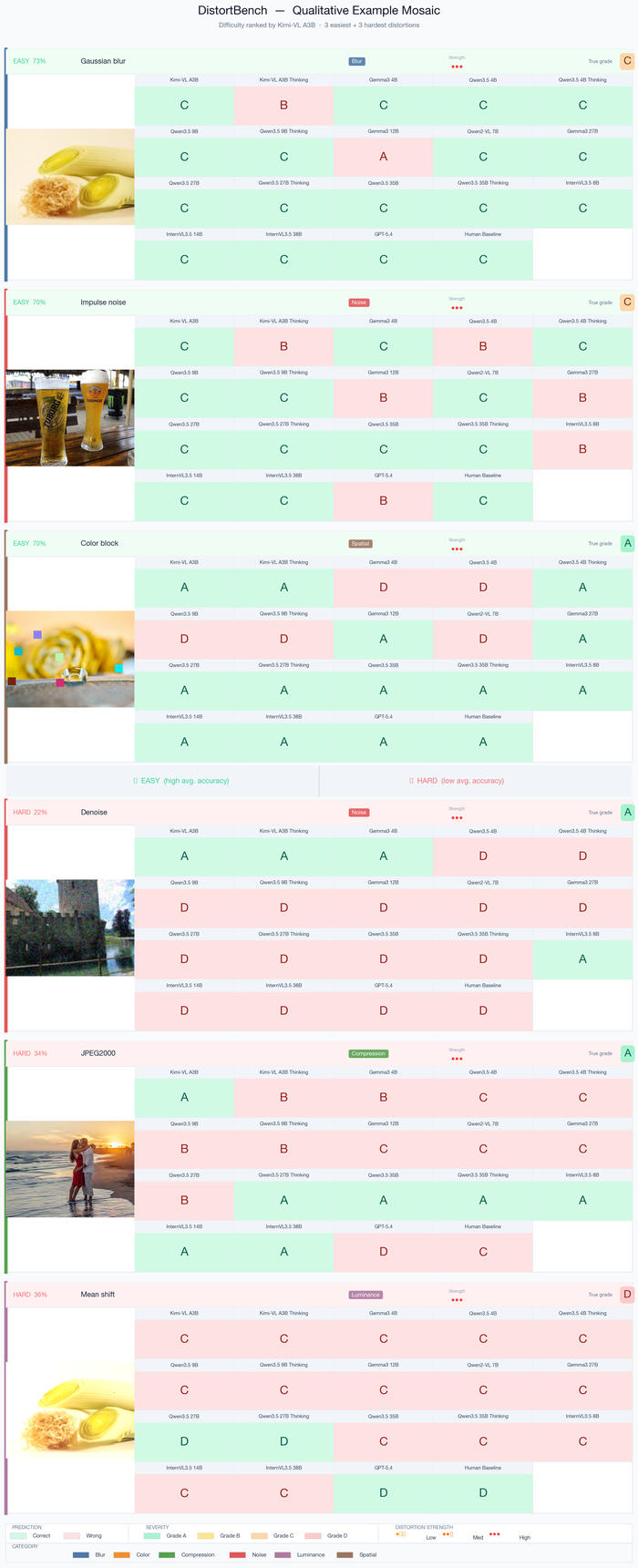}
    \caption{\textbf{Qualitative examples.} Six distortion--level pairs showing the distorted image alongside per-model predictions. Correct predictions are marked in \textcolor{green!60!black}{green}, incorrect in \textcolor{red}{red}. The mosaic spans both easy cases (rotation, Gaussian blur) where most models succeed and hard cases (non-eccentricity, jitter) where most fail.}
    \label{fig:mosaic}
\end{figure}

\cref{fig:mosaic} presents qualitative examples spanning the difficulty spectrum.
On easy distortions such as rotation and Gaussian blur at high severity, all five models predict correctly, confirming that these distortions produce salient visual signatures.
On hard distortions such as non-eccentricity and jitter, even the best models fail, often confusing them with visually similar distortions.
The mosaic also reveals that severity level~1 examples remain challenging despite relatively strong aggregate accuracy for some models: models frequently select a distractor with the correct type but wrong severity, consistent with the severity bias observed in the paper.

\appsectionspace
\section{Distortion Generation Parameters}
\label{app:params}

\cref{tab:distortion_params} lists the exact parameter values used to generate each severity level for all 27 distortion types.

For the 25 KADID-10k distortions, these values were manually calibrated by Lin~\emph{et al.}~\cite{lin2019kadid} so that visual quality varies perceptually roughly linearly across the five levels, and were validated via a crowdsourced DCR study with 30 ratings per image. 
The two rotation distortions (types 26--27) use author-defined angular increments.

\begin{table}[!t]
\centering
\caption{\textbf{Distortion generation parameters.} Exact parameter values for each severity level (L1=Very mild to L5=Very strong) across all 27 distortion types. Parameter semantics vary by distortion (e.g., $\sigma$ for Gaussian blur, quality factor for JPEG). Higher levels produce stronger degradation in all cases.}
\label{tab:distortion_params}
\footnotesize
\setlength{\tabcolsep}{3pt}
\resizebox{\linewidth}{!}{%
\begin{tabular}{@{}rlccccc@{}}
\toprule
\textbf{ID} & \textbf{Distortion} & \textbf{L1} & \textbf{L2} & \textbf{L3} & \textbf{L4} & \textbf{L5} \\
\midrule
1  & Gaussian blur          & 0.1   & 0.5   & 1     & 2     & 5     \\
2  & Lens blur              & 1     & 2     & 4     & 6     & 8     \\
3  & Motion blur            & 1     & 2     & 4     & 6     & 10    \\
4  & Color diffusion        & 1     & 3     & 6     & 8     & 12    \\
5  & Color shift            & 1     & 3     & 6     & 8     & 12    \\
6  & Color quantization     & 64    & 48    & 32    & 16    & 8     \\
7  & Color saturation (over)& 1.2   & 1.5   & 2.0   & 3.0   & 5.0   \\
8  & Color saturation (de)  & 1     & 2     & 3     & 6     & 9     \\
9  & JPEG2000               & 16    & 32    & 45    & 120   & 400   \\
10 & JPEG                   & 43    & 36    & 24    & 7     & 4     \\
11 & White noise            & 0.001 & 0.002 & 0.003 & 0.005 & 0.01  \\
12 & White noise (color)    & 0.0001& 0.0005& 0.001 & 0.002 & 0.003 \\
13 & Impulse noise          & 0.001 & 0.005 & 0.01  & 0.02  & 0.03  \\
14 & Multiplicative noise   & 0.001 & 0.005 & 0.01  & 0.02  & 0.05  \\
15 & Denoise                & 0.01  & 0.03  & 0.05  & 0.1   & 0.15  \\
16 & Brighten               & 0.1   & 0.2   & 0.4   & 0.7   & 1.1   \\
17 & Darken                 & 0.05  & 0.1   & 0.2   & 0.4   & 0.8   \\
18 & Mean shift             & 0.05  & 0.1   & 0.15  & 0.2   & 0.25  \\
19 & Jitter                 & 0.05  & 0.1   & 0.2   & 0.5   & 1     \\
20 & Non-eccentricity       & 20    & 40    & 60    & 80    & 100   \\
21 & Pixelate               & 0.01  & 0.05  & 0.1   & 0.2   & 0.5   \\
22 & Quantization           & 20    & 16    & 13    & 10    & 7     \\
23 & Color block            & 2     & 4     & 6     & 8     & 10    \\
24 & High sharpen           & 1     & 2     & 3     & 6     & 12    \\
25 & Contrast change        & 0.1   & 0.2   & 0.35  & 0.5   & 0.7   \\
26 & Rotation (CW)          & $2^\circ$ & $5^\circ$ & $10^\circ$ & $20^\circ$ & $45^\circ$ \\
27 & Rotation (CCW)         & $2^\circ$ & $5^\circ$ & $10^\circ$ & $20^\circ$ & $45^\circ$ \\
\bottomrule
\end{tabular}
}
\end{table}

\clearpage
\section{Overall Metrics Comparison}
\label{app:metrics}

\begin{figure}[H]
    \centering
    \includegraphics[width=\linewidth]{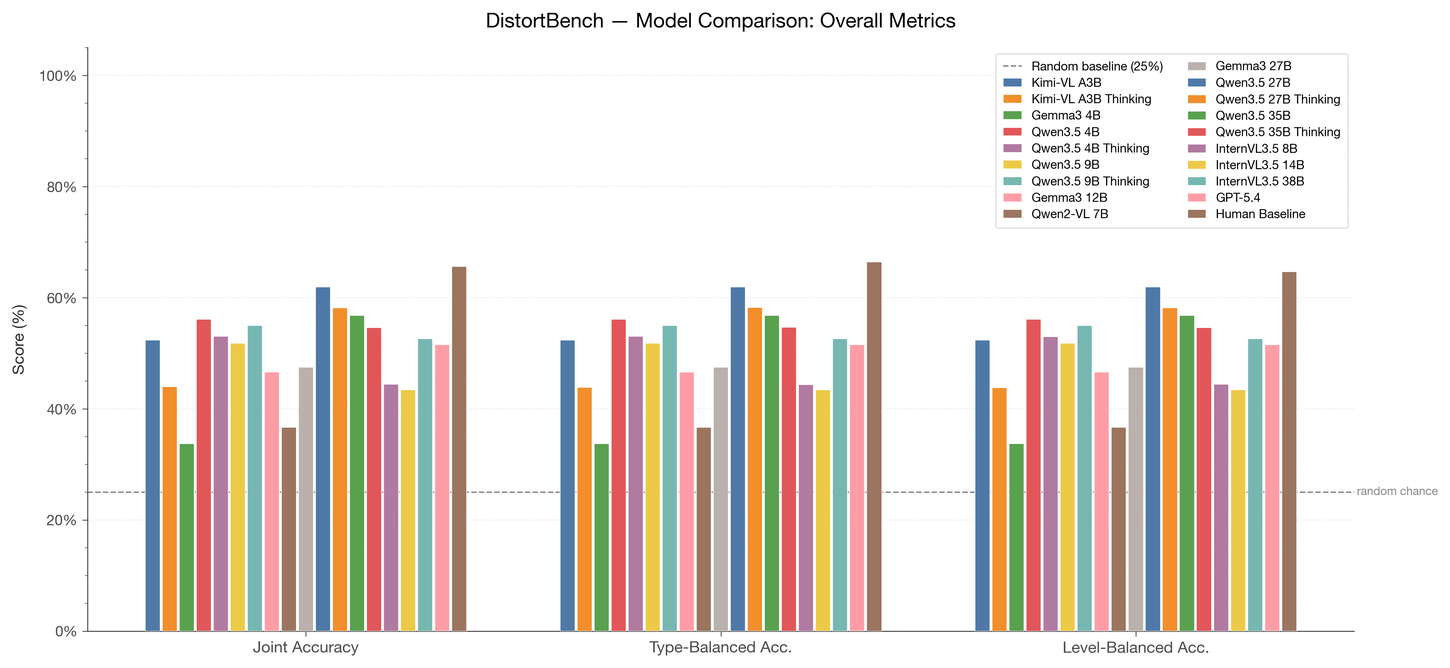}
    \caption{\textbf{Overall metrics comparison.} Grouped bar chart showing joint accuracy, type-balanced accuracy, and level-balanced accuracy for all models. The dashed line marks the 25\% random baseline. Qwen3.5 base models consistently lead; thinking variants underperform their base counterparts in four of five pairs (the exception being Qwen3.5~9B Think., which improves over base by 3.2~pp).}
    \label{fig:multi_metric_bar}
\end{figure}

\Cref{fig:multi_metric_bar} provides a direct visual comparison of the three headline metrics reported in the paper: joint accuracy, type-balanced accuracy, and level-balanced accuracy.
Because DistortBench is exactly balanced across distortion types and severity levels, the three bars are nearly identical for every model, confirming that the overall ranking is not being driven by class imbalance.
The figure therefore mainly serves as a compact visual summary of the leaderboard while making it easy to compare base and thinking variants side by side.

\clearpage
\section{KADID Crowd-Rating Sanity Check}
\label{app:kadid_human}

To further validate that the inherited KADID severity levels align with human perception, we aggregated the raw crowd-rating records in the released KADID metadata for the 10{,}125 \texttt{kon10k\_png} distorted images (excluding the auxiliary \texttt{tid\_2013\_png} entries present in the same CSV dump).
Each distorted image has a median of 30 human ratings.
Using the mean per-image human quality score, the inherited five-level severity scale shows a clear monotonic degradation trend overall: average human score drops from 4.08 at level~1 to 2.01 at level~5.
Moreover, 22 of the 25 inherited distortion types exhibit monotonic non-increasing mean human scores across levels.
The three exceptions are color shift, which shows only a very small level~4$\to$5 uptick, and high sharpen and contrast change, which are rated more favorably at mild-to-moderate levels before declining at stronger settings.
This analysis does not provide human accuracy on DistortBench's MCQ task, but it offers an additional human-perception sanity check for the severity calibration used by the benchmark.

\begin{table}[H]
\centering
\caption{\textbf{KADID crowd-rating sanity check.} Aggregate human quality scores from the raw KADID crowd-rating metadata for the 25 inherited distortion types. Higher scores indicate better perceptual quality, so monotonic decline with level supports the intended severity ordering.}
\label{tab:kadid_human}
\small
\setlength{\tabcolsep}{5pt}
\begin{tabular}{@{}cccc@{}}
\toprule
\textbf{Level} & \textbf{Mean score} & \textbf{Std. across images} & \textbf{\# images} \\
\midrule
1 & 4.08 & 0.61 & 2025 \\
2 & 3.52 & 0.81 & 2025 \\
3 & 3.06 & 0.90 & 2025 \\
4 & 2.50 & 0.87 & 2025 \\
5 & 2.01 & 0.77 & 2025 \\
\bottomrule
\end{tabular}
\end{table}

\section{Human Annotation Study Design}

\begin{figure}[H]
    \centering
    \includegraphics[width=\linewidth,height=0.5\textheight,keepaspectratio]{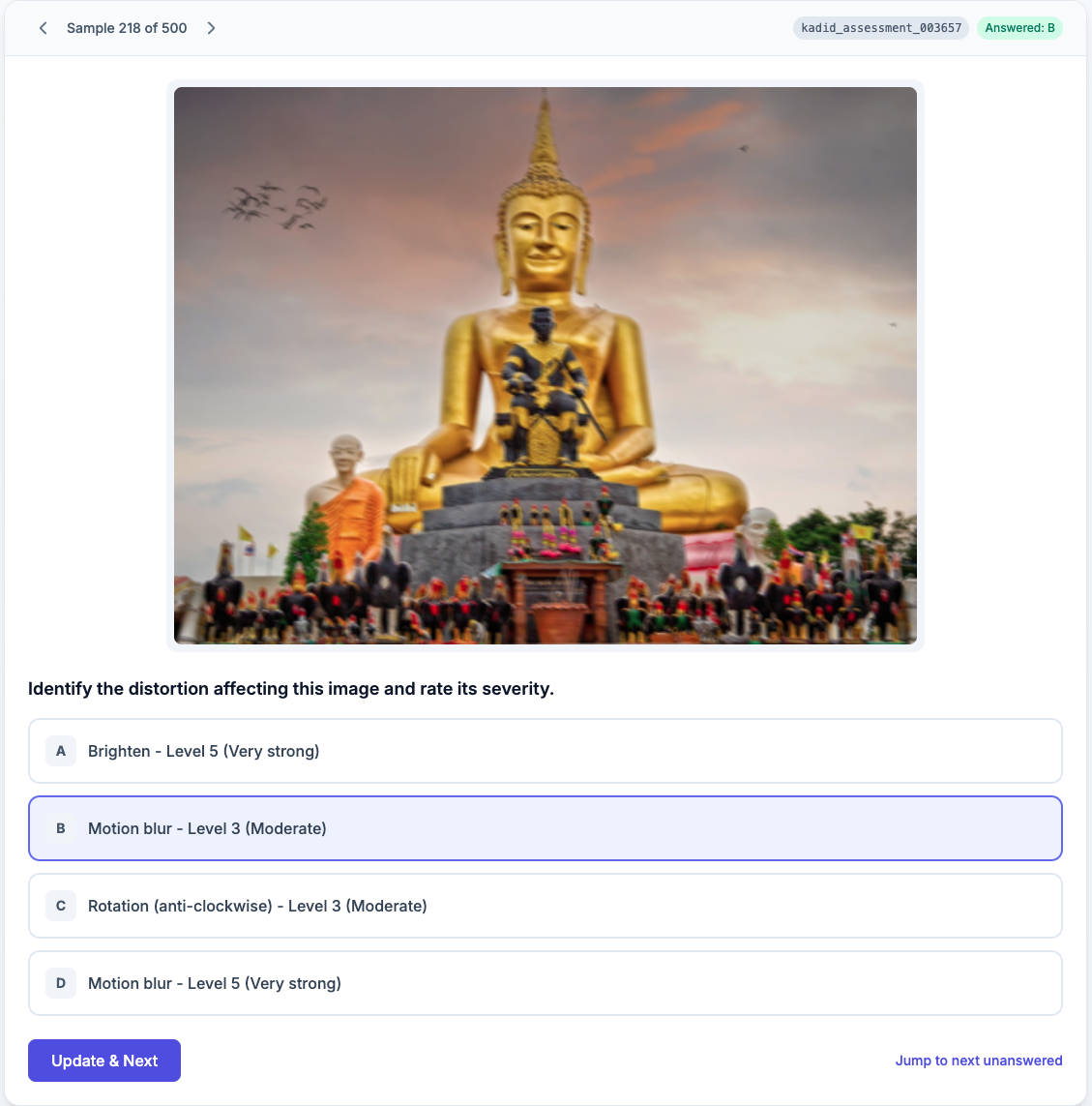}
    \caption{\textbf{Snapshot showing the human annotation study design.} }
    \label{fig:human_annotation_study}
\end{figure}

%% file: main.bib
@article{wang2004image,
  title={Image quality assessment: from error visibility to structural similarity},
  author={Wang, Zhou and Bovik, Alan C and Sheikh, Hamid R and Simoncelli, Eero P},
  journal={IEEE Transactions on Image Processing},
  volume={13},
  number={4},
  pages={600--612},
  year={2004},
  publisher={IEEE}
}

@inproceedings{zhang2018unreasonable,
  title={The unreasonable effectiveness of deep features as a perceptual metric},
  author={Zhang, Richard and Isola, Phillip and Efros, Alexei A and Shechtman, Eli and Wang, Oliver},
  booktitle={CVPR},
  pages={586--595},
  year={2018}
}

@article{mittal2012no,
  title={No-reference image quality assessment in the spatial domain},
  author={Mittal, Anish and Moorthy, Anush Krishna and Bovik, Alan Conrad},
  journal={IEEE Transactions on Image Processing},
  volume={21},
  number={12},
  pages={4695--4708},
  year={2012},
  publisher={IEEE}
}

@inproceedings{ke2021musiq,
  title={{MUSIQ}: Multi-scale image quality transformer},
  author={Ke, Junjie and Wang, Qifei and Wang, Yilin and Milanfar, Peyman and Yang, Feng},
  booktitle={ICCV},
  pages={5148--5157},
  year={2021}
}

@inproceedings{su2020blindly,
  title={Blindly assess image quality in the wild guided by a self-adaptive hyper network},
  author={Su, Shaolin and Yan, Qingsen and Zhu, Yu and Zhang, Cheng and Ge, Xin and Sun, Jinqiu and Zhang, Yanning},
  booktitle={CVPR},
  pages={3667--3676},
  year={2020}
}

@article{sheikh2006statistical,
  title={A statistical evaluation of recent full reference image quality assessment algorithms},
  author={Sheikh, Hamid R and Sabir, Muhammad F and Bovik, Alan C},
  journal={IEEE Transactions on Image Processing},
  volume={15},
  number={11},
  pages={3440--3451},
  year={2006},
  publisher={IEEE}
}

@article{larson2010most,
  title={Most apparent distortion: full-reference image quality assessment and the role of strategy},
  author={Larson, Eric Cooper and Chandler, Damon Michael},
  journal={Journal of Electronic Imaging},
  volume={19},
  number={1},
  pages={011006},
  year={2010},
  publisher={SPIE}
}

@article{ponomarenko2015image,
  title={Image database {TID2013}: Peculiarities, results and perspectives},
  author={Ponomarenko, Nikolay and Jin, Lina and Ieremeiev, Oleg and Lukin, Vladimir and Egiazarian, Karen and Astola, Jaakko and Vozel, Benoit and Chehdi, Kacem and Carli, Marco and Battisti, Federica and others},
  journal={Signal Processing: Image Communication},
  volume={30},
  pages={57--77},
  year={2015},
  publisher={Elsevier}
}

@inproceedings{lin2019kadid,
  title={{KADID-10K}: A large-scale artificially distorted {IQA} database},
  author={Lin, Hanhe and Hosu, Vlad and Saupe, Dietmar},
  booktitle={International Conference on Quality of Multimedia Experience (QoMEX)},
  pages={1--3},
  year={2019},
  publisher={IEEE}
}

@inproceedings{liu2023visual,
  title={Visual instruction tuning},
  author={Liu, Haotian and Li, Chunyuan and Wu, Qingyang and Lee, Yong Jae},
  booktitle={NeurIPS},
  year={2023}
}

@misc{team2025kimi,
      title={Kimi-VL Technical Report}, 
      author={Kimi Team and Angang Du and Bohong Yin and Bowei Xing and Bowen Qu and Bowen Wang and Cheng Chen and Chenlin Zhang and Chenzhuang Du and Chu Wei and Congcong Wang and Dehao Zhang and Dikang Du and Dongliang Wang and Enming Yuan and Enzhe Lu and Fang Li and Flood Sung and Guangda Wei and Guokun Lai and Han Zhu and Hao Ding and Hao Hu and Hao Yang and Hao Zhang and Haoning Wu and Haotian Yao and Haoyu Lu and Heng Wang and Hongcheng Gao and Huabin Zheng and Jiaming Li and Jianlin Su and Jianzhou Wang and Jiaqi Deng and Jiezhong Qiu and Jin Xie and Jinhong Wang and Jingyuan Liu and Junjie Yan and Kun Ouyang and Liang Chen and Lin Sui and Longhui Yu and Mengfan Dong and Mengnan Dong and Nuo Xu and Pengyu Cheng and Qizheng Gu and Runjie Zhou and Shaowei Liu and Sihan Cao and Tao Yu and Tianhui Song and Tongtong Bai and Wei Song and Weiran He and Weixiao Huang and Weixin Xu and Xiaokun Yuan and Xingcheng Yao and Xingzhe Wu and Xinhao Li and Xinxing Zu and Xinyu Zhou and Xinyuan Wang and Y. Charles and Yan Zhong and Yang Li and Yangyang Hu and Yanru Chen and Yejie Wang and Yibo Liu and Yibo Miao and Yidao Qin and Yimin Chen and Yiping Bao and Yiqin Wang and Yongsheng Kang and Yuanxin Liu and Yuhao Dong and Yulun Du and Yuxin Wu and Yuzhi Wang and Yuzi Yan and Zaida Zhou and Zhaowei Li and Zhejun Jiang and Zheng Zhang and Zhilin Yang and Zhiqi Huang and Zihao Huang and Zijia Zhao and Ziwei Chen and Zongyu Lin},
      year={2025},
      eprint={2504.07491},
      archivePrefix={arXiv},
      primaryClass={cs.CV},
      url={https://arxiv.org/abs/2504.07491}, 
}

@article{wang2024qwen2,
  title={Qwen2-{VL}: Enhancing vision-language model's perception of the world at any resolution},
  author={Wang, Peng and Bai, Shuai and Tan, Sinan and Wang, Shijie and Fan, Zhihao and Bai, Jinze and Chen, Keqin and Liu, Xuejing and Wang, Jialin and Ge, Wenbin and others},
  journal={arXiv preprint arXiv:2409.12191},
  year={2024}
}

@inproceedings{wu2024qbench,
  title={{Q-Bench}: A benchmark for general-purpose foundation models on low-level vision},
  author={Wu, Haoning and Zhang, Zicheng and Zhang, Erli and Chen, Chaofeng and Liao, Liang and Wang, Annan and Li, Chunyi and Sun, Wenxiu and Yan, Qiong and Zhai, Guangtao and others},
  booktitle={ICLR},
  year={2024}
}

@article{wu2023qalign,
  title={{Q-Align}: Teaching {LMMs} for visual scoring via discrete text-defined levels},
  author={Wu, Haoning and Zhang, Zicheng and Zhang, Erli and Chen, Chaofeng and Liao, Liang and Wang, Annan and Xu, Kaixin and Li, Chunyi and Hou, Jingwen and Zhai, Guangtao and others},
  journal={ICML},
  year={2024}
}

@article{you2024depicting,
  title={Depicting beyond scores: Advancing image quality assessment through multi-modal language models},
  author={You, Zhiyuan and Li, Zheyuan and Gu, Jinjin and Yin, Zhenfei and Xue, Tianfan and Dong, Chao},
  journal={arXiv preprint arXiv:2312.08962},
  year={2024}
}

@inproceedings{liu2023mmbench,
  title={{MMBench}: Is your multi-modal model an all-around player?},
  author={Liu, Yuan and Duan, Haodong and Zhang, Yuanhan and Li, Bo and Zhang, Songyang and Zhao, Wangbo and Yuan, Yike and Wang, Jiaqi and He, Conghui and Liu, Ziwei and others},
  booktitle={ECCV},
  year={2024}
}

@inproceedings{yue2024mmmu,
  title={{MMMU}: A massive multi-discipline multimodal understanding and reasoning benchmark for expert {AGI}},
  author={Yue, Xiang and Ni, Yuansheng and Zhang, Kai and Zheng, Tianyu and Liu, Ruoqi and Zhang, Ge and Stevens, Samuel and Jiang, Dongfu and Ren, Weiming and Sun, Yuxuan and others},
  booktitle={CVPR},
  year={2024}
}

@inproceedings{lu2022learn,
  title={Learn to explain: Multimodal reasoning via thought chains for science question answering},
  author={Lu, Pan and Mishra, Swaroop and Xia, Tony and Qiu, Liang and Chang, Kai-Wei and Zhu, Song-Chun and Tafjord, Oyvind and Clark, Peter and Kalyan, Ashwin},
  booktitle={NeurIPS},
  year={2022}
}

@inproceedings{kwon2023efficient,
  title={Efficient memory management for large language model serving with {PagedAttention}},
  author={Kwon, Woosuk and Li, Zhuohan and Zhuang, Siyuan and Sheng, Ying and Liang, Yinmin and Stoica, Ion and Zheng, Lianmin},
  booktitle={SOSP},
  year={2023}
}

@article{wei2022chain,
  title={Chain-of-thought prompting elicits reasoning in large language models},
  author={Wei, Jason and Wang, Xuezhi and Schuurmans, Dale and Bosma, Maarten and Ichter, Brian and Xia, Fei and Chi, Ed and Le, Quoc V and Zhou, Denny},
  journal={NeurIPS},
  year={2022}
}

@article{bosse2018deep,
  title={Deep neural networks for no-reference and full-reference image quality assessment},
  author={Bosse, Sebastian and Maniry, Dominique and M{\"u}ller, Klaus-Robert and Wiegand, Thomas and Samek, Wojciech},
  journal={IEEE Transactions on Image Processing},
  volume={27},
  number={1},
  pages={206--219},
  year={2018},
  publisher={IEEE}
}

@inproceedings{zhang2021designing,
  title={Designing a practical degradation model for deep blind image super-resolution},
  author={Zhang, Kai and Liang, Jingyun and Van Gool, Luc and Timofte, Radu},
  booktitle={ICCV},
  pages={4791--4800},
  year={2021}
}

@article{verdoliva2020media,
  title={Media forensics and {DeepFakes}: An overview},
  author={Verdoliva, Luisa},
  journal={IEEE Journal of Selected Topics in Signal Processing},
  volume={14},
  number={5},
  pages={910--932},
  year={2020},
  publisher={IEEE}
}

@article{zhang2024benchmark,
  title={A benchmark for multi-modal foundation models on low-level vision: from single images to pairs},
  author={Zhang, Zicheng and Wu, Haoning and Zhang, Erli and Zhai, Guangtao and Min, Xiongkuo},
  journal={arXiv preprint arXiv:2402.07116},
  year={2024}
}

@misc{openai2025gpt5,
  title={Introducing {GPT-5.4}},
  author={{OpenAI}},
  howpublished={\url{https://openai.com/index/introducing-gpt-5-4/}},
  year={2026}
}

@article{bai2025qwen25vl,
  title={Qwen2.5-{VL} Technical Report},
  author={Bai, Shuai and Chen, Keqin and Liu, Xuejing and Wang, Jialin and Ge, Wenbin and Song, Sibo and Dang, Kai and Wang, Peng and Wang, Shijie and Tang, Jun and others},
  journal={arXiv preprint arXiv:2502.13923},
  year={2025}
}

@article{team2025gemma3,
  title={Gemma 3 Technical Report},
  author={{Gemma Team}},
  journal={arXiv preprint arXiv:2503.19786},
  year={2025}
}

@article{chen2024internvl25,
  title={Expanding Performance Boundaries of Open-Source Multimodal Models with Model, Data, and Test-Time Scaling},
  author={Chen, Zhe and Wang, Weiyun and Cao, Yue and Liu, Yangzhou and Gao, Zhangwei and Cui, Erfei and Zhu, Jinguo and Ye, Shenglong and Tian, Hao and Liu, Zhaoyang and others},
  journal={arXiv preprint arXiv:2412.05271},
  year={2024}
}
